\documentclass{article}
\usepackage{spconf,amsmath,graphicx}
\usepackage{booktabs,multirow}
\usepackage[caption=false,font=footnotesize]{subfig}
\usepackage{url}
\usepackage{hyperref}

\title{How fragile are training-free AI-generated image detectors?\\
A controlled audit of score direction, preprocessing, and compression}

\name{Jingwen Zhou \qquad Mingzhe Wang}
\address{Xidian University, Xi'an, China}

\begin{document}
\ninept
\maketitle

\begin{abstract}
Training-free detectors of AI-generated images promise generator-agnostic
deployment without classifier training, yet their reported numbers are rarely
compared under a single controlled protocol. We audit two representative
training-free scores---an autoencoder-reconstruction score (AEROBLADE-style) and a
noise-perturbation feature-similarity score (RIGID-style)---plus a na\"ive
feature-kNN control, on a common 1{,}500-image GenImage-derived benchmark
spanning seven generators and JPEG compression at quality 70 and 50. The audit
yields three cautionary findings. (i)~\emph{Implementation details masquerade as
method differences}: replacing the LPIPS backbone (AlexNet$\to$VGG-16) changes
overall AUROC by $+0.085$, and switching between resize-to-512 and
native-resolution preprocessing flips per-generator conclusions by up to $0.38$
AUROC. (ii)~\emph{Score direction is not a property of the method but of its
hyperparameters}: the RIGID-style score is \emph{inverted} (AUROC $<0.5$) on SD1.5
and Wukong at noise level $\sigma{=}0.05$, recovers to $>0.5$ for every generator
at $\sigma{=}0.01$, and collapses to $0.15$ at $\sigma{=}0.3$.
(iii)~\emph{Dataset format bias inflates robustness claims}: without unified
re-encoding, AUROC under JPEG-50 \emph{exceeds} the clean condition for the
AlexNet-backbone reconstruction score; after bias correction the residual
anomaly localizes to a single generator (BigGAN). The audited scores have complementary per-generator failure
sets, but na\"ive z-score fusion does not beat the best single score, indicating
that exploiting complementarity requires direction-aware combination.
\end{abstract}


\section{Introduction}
\label{sec:intro}

The rapid progress of generative models---from
GANs~\cite{goodfellow2014gan,brock2019biggan} to diffusion
models~\cite{ho2020ddpm,dhariwal2021adm} and text-conditional systems such as
GLIDE, latent diffusion, VQ-diffusion, and
SDXL~\cite{nichol2022glide,rombach2022ldm,gu2022vqdiffusion,podell2024sdxl}---has
made synthetic imagery a first-order forensic concern. The classical response is
\emph{training-based}: convolutional classifiers trained on real/fake pairs
generalize surprisingly well with suitable augmentation~\cite{wang2020cnn}, but
degrade under compression, resizing, and unseen
architectures~\cite{gragnaniello2021gan,corvi2023icassp}. A second family
exploits generator fingerprints and the frequency artifacts left by
upsampling~\cite{marra2019fingerprints,frank2020frequency}, artifacts that
persist, in attenuated form, in modern diffusion
models~\cite{corvi2023intriguing,tan2024npr}. A third line builds on large
pretrained representations such as CLIP~\cite{radford2021clip}, via
nearest-neighbor or linear probing~\cite{ojha2023universal} or mixtures of low-
and high-level experts~\cite{yan2025aide}.

Against this backdrop, detection is increasingly framed as a
\emph{training-free} problem: rather than training a classifier at all, one
computes a score from a frozen pretrained model---e.g., the reconstruction error
of a latent-diffusion autoencoder~\cite{ricker2024aeroblade}, the sensitivity of
self-supervised features to noise perturbations~\cite{he2024rigid}, diffusion
reconstruction error~\cite{wang2023dire}, high-frequency aliasing of the LDM
autoencoder~\cite{choi2024hfi}, edit-induced reconstruction
shifts~\cite{jiang2025eires}, or the coding cost of an image under a lossless
model of real images~\cite{cozzolino2024zed}---and thresholds it. Recent
refinements combine multiple perturbation types~\cite{tsai2024minder}.
Training-free scores are attractive because they promise generalization to
unseen generators and trivial deployment.

However, the literature reports these scores under heterogeneous protocols:
different real/fake sources, image resolutions and crop policies,
perceptual-metric implementations, and (often unreported) robustness pipelines.
Recent work has shown that \emph{trained} detectors are heavily confounded by
dataset format biases such as the JPEG provenance of real
images~\cite{grommelt2024fakeorjpeg,guillaro2025bfree}, and concurrent work
observes that the perturbation-robustness assumption behind RIGID-like scores
admits exceptions and benefits from searching layer/perturbation
configurations~\cite{huang2026intermediate}. A systematic, single-protocol
quantification of how fragile training-free scores are to these factors is still
missing.

This paper contributes a deliberately small, fully controlled \emph{audit} rather
than a new detector. We re-implement two representative training-free scores plus
a na\"ive control, freeze a single evaluation set, and vary one factor at a time:
perceptual-metric backbone, preprocessing resolution policy, perturbation strength
and feature depth, and JPEG re-encoding of the dataset itself. Our findings
quantify how sensitive
published training-free numbers can be to choices that are usually relegated to
appendices, and motivate minimal reporting standards for this rapidly growing
subfield.

\section{Audited methods and protocol}
\label{sec:methods}

\subsection{Benchmark}
We sample 1{,}500 images from the validation portion of a
GenImage~\cite{zhu2023genimage} repackaging\footnote{\texttt{TheKernel01/%
Tiny-GenImage} on Hugging Face; images originate from GenImage. The validation
split contains seven of the eight GenImage generators (no SD1.4).}: 800 real
images (ImageNet) and 700 fakes, 100 each from seven
generators (ADM~\cite{dhariwal2021adm}, BigGAN~\cite{brock2019biggan},
GLIDE~\cite{nichol2022glide}, Midjourney, SD1.5~\cite{rombach2022ldm},
VQDM~\cite{gu2022vqdiffusion}, Wukong), drawn with a fixed seed (42). We report
threshold-free AUROC with fake as the positive class; per-generator cells use that
generator's 100 fakes against all 800 reals. Compression robustness is measured
by re-encoding every image at JPEG quality $q{=}70$ and $q{=}50$ (Pillow).
All experiments run on a single A100; no component is trained or tuned on the
evaluation data.

\subsection{Score 1: autoencoder reconstruction (AEROBLADE-style)}
Following AEROBLADE~\cite{ricker2024aeroblade}, an image $x$ is encoded and
decoded by the Stable Diffusion VAE~\cite{rombach2022ldm}
(\texttt{sd-vae-ft-mse}), and the score is the perceptual distance
$d_{\mathrm{LPIPS}}(x,\hat x)$~\cite{zhang2018lpips}; fakes reconstruct better
(lower distance). We audit two factors: \textbf{(a)~LPIPS backbone}:
AlexNet~\cite{krizhevsky2012alexnet} vs.\ VGG-16~\cite{simonyan2015vgg} (the
original paper uses VGG); \textbf{(b)~preprocessing}: bicubic short-side resize
to 512 + center crop (\emph{resize-512}) vs.\ native-resolution center crop to a
multiple of 16, capped at 512 (\emph{native}).

\subsection{Score 2: perturbation sensitivity (RIGID-style)}
Following RIGID~\cite{he2024rigid}, we compute
DINOv2-base~\cite{oquab2024dinov2} embeddings of $x$ and of $x+\varepsilon$,
$\varepsilon\sim\mathcal N(0,\sigma^2 I)$, and use their cosine similarity
(averaged over 3 noise draws, fixed seed) as the score; real images are assumed
more robust (higher similarity). Inputs are bicubically resized and center-cropped
to $224^2$. We grid $\sigma\in\{0.01,0.05,0.1,0.3\}$ (pixel scale, $[0,1]$
images) $\times$ feature depth \{layer-6 CLS (\emph{mid}), final CLS
(\emph{last})\}.

\subsection{Score 3: feature-kNN control}
As a non-reconstruction, non-perturbation control (cf.\ nearest-neighbor
classification in CLIP space~\cite{ojha2023universal}), each image is scored by
its mean cosine distance to the $k{=}5$ nearest neighbors among 1{,}000
\emph{held-out} real images (disjoint from the test reals; DINOv2 final CLS
features). Larger distance $\Rightarrow$ fake.

\subsection{Format-bias correction}
In GenImage, real images are JPEG-sourced while fakes are PNG---a known
confound~\cite{grommelt2024fakeorjpeg}. Our \emph{bias-corrected} pipeline
therefore re-encodes all images at JPEG $q{=}95$ (condition \texttt{q95}) and
applies the $q{=}70/50$ degradations on top, so real and fake share identical
final compression provenance.

\section{Results}
\label{sec:results}

\subsection{Backbone and preprocessing dominate the headline number}
\begin{table}[t]
\centering
\caption{Overall AUROC, original pipeline (800 real / 700 fake; fake is the
positive class). The LPIPS backbone swap alone is worth $+0.085$ AUROC; the
preprocessing policy changes the direction of the compression trend.}
\label{tab:overall}
\footnotesize
\setlength{\tabcolsep}{4.5pt}
\begin{tabular}{llccc}
\toprule
Score & Variant & clean & JPEG-70 & JPEG-50 \\
\midrule
\multirow{4}{*}{AEROBLADE-style}
 & alex, resize-512 & 0.740 & 0.757 & 0.762 \\
 & \textbf{vgg, resize-512} & \textbf{0.825} & 0.794 & 0.776 \\
 & alex, native & 0.735 & 0.708 & 0.693 \\
 & vgg, native & 0.785 & 0.725 & 0.694 \\
\midrule
RIGID-style & $\sigma{=}0.05$, last & 0.692 & 0.673 & 0.677 \\
kNN control & $k{=}5$, last & 0.568 & 0.558 & 0.557 \\
\midrule
Fusion (z-score) & alex-512$+$RIGID & 0.762 & 0.754 & 0.758 \\
\bottomrule
\end{tabular}
\end{table}

Table~\ref{tab:overall} shows the original-pipeline matrix. Two observations.
First, swapping the LPIPS backbone from AlexNet to VGG-16---a one-line
change---moves overall clean AUROC from $0.740$ to $0.825$. Any leaderboard that
mixes implementations is thus comparing backbones as much as methods. Second, the
\emph{direction of the compression trend} depends on preprocessing: with
resize-512 the AlexNet score \emph{improves} under JPEG ($0.740\to0.762$), while
native preprocessing degrades monotonically ($0.735\to0.693$).
Sec.~\ref{ssec:bias} traces the non-monotonicity to dataset format bias plus a
single generator. The kNN control stays near chance ($\approx 0.56$) in all
conditions, confirming that the reconstruction and perturbation scores carry
signal beyond plain feature-space proximity to real images.

\begin{table}[t]
\centering
\caption{Per-generator AUROC (clean, original pipeline): each generator's 100
fakes vs.\ all 800 reals. Bold marks the largest preprocessing-induced swings
(BigGAN, Midjourney) and below-chance cells (RIGID on SD1.5/Wukong). The two
score families have complementary failure sets.}
\label{tab:pergen}
\footnotesize
\setlength{\tabcolsep}{4pt}
\begin{tabular}{lccccc}
\toprule
Generator & \begin{tabular}{@{}c@{}}AERO\\alex-512\end{tabular}
          & \begin{tabular}{@{}c@{}}AERO\\alex-nat.\end{tabular}
          & \begin{tabular}{@{}c@{}}AERO\\vgg-512\end{tabular}
          & \begin{tabular}{@{}c@{}}AERO\\vgg-nat.\end{tabular}
          & \begin{tabular}{@{}c@{}}RIGID\\$\sigma{=}.05$\end{tabular} \\
\midrule
ADM        & 0.714 & 0.663 & 0.772 & 0.697 & 0.816 \\
BigGAN     & 0.812 & 0.605 & \textbf{0.911} & \textbf{0.528} & 0.931 \\
GLIDE      & 0.956 & 0.885 & 0.983 & 0.922 & 0.822 \\
Midjourney & 0.804 & 0.954 & \textbf{0.834} & \textbf{0.989} & 0.587 \\
SD1.5      & 0.647 & 0.706 & 0.807 & 0.840 & \textbf{0.408} \\
VQDM       & 0.626 & 0.650 & 0.649 & 0.664 & 0.834 \\
Wukong     & 0.624 & 0.685 & 0.822 & 0.853 & \textbf{0.445} \\
\bottomrule
\end{tabular}
\end{table}

Table~\ref{tab:pergen} breaks results down by generator. The two families are
strikingly complementary: RIGID-style is strongest exactly where reconstruction
is weakest (ADM, BigGAN, VQDM) and vice versa (GLIDE, Midjourney). Preprocessing
alone swings BigGAN by $0.38$ AUROC under the VGG backbone ($0.911$ resize-512
vs.\ $0.528$ native) and Midjourney by $0.155$ in the \emph{opposite} direction
($0.834$ vs.\ $0.989$). A cross-generator ranking computed under one
preprocessing policy does not survive the other.

\subsection{Score direction is hyperparameter-contingent}
\begin{table}[t]
\centering
\caption{RIGID-style AUROC vs.\ noise level $\sigma$ and feature depth (clean
images, original pipeline; MJ $=$ Midjourney, Wu $=$ Wukong). Bold: inverted
direction (AUROC $<0.5$). The SD1.5/Wukong inversion at $\sigma{=}0.05$
disappears at $\sigma{=}0.01$ and deepens catastrophically at $\sigma{=}0.3$;
VQDM inverts with \emph{mid} features but not with \emph{last}.}
\label{tab:rigidgrid}
\scriptsize
\setlength{\tabcolsep}{2.6pt}
\begin{tabular}{llcccccccc}
\toprule
$\sigma$ & layer & ALL & ADM & BigGAN & GLIDE & MJ & SD1.5 & VQDM & Wu \\
\midrule
0.01 & mid  & 0.759 & 0.782 & 0.911 & 0.964 & 0.865 & 0.620 & 0.571 & 0.601 \\
0.01 & last & 0.793 & 0.848 & 0.967 & 0.973 & 0.787 & 0.534 & 0.843 & 0.598 \\
0.05 & mid  & 0.678 & 0.763 & 0.886 & 0.861 & 0.841 & 0.517 & \textbf{0.480} & \textbf{0.402} \\
0.05 & last & 0.691 & 0.815 & 0.933 & 0.823 & 0.589 & \textbf{0.405} & 0.831 & \textbf{0.444} \\
0.10 & mid  & 0.611 & 0.743 & 0.774 & 0.790 & 0.795 & \textbf{0.442} & \textbf{0.418} & \textbf{0.313} \\
0.10 & last & 0.614 & 0.764 & 0.851 & 0.689 & 0.500 & \textbf{0.336} & 0.790 & \textbf{0.371} \\
0.30 & mid  & 0.571 & 0.692 & 0.755 & 0.732 & 0.691 & \textbf{0.402} & \textbf{0.414} & \textbf{0.310} \\
0.30 & last & \textbf{0.361} & 0.553 & \textbf{0.446} & \textbf{0.394} & \textbf{0.341} & \textbf{0.149} & \textbf{0.474} & \textbf{0.169} \\
\bottomrule
\end{tabular}
\end{table}

\begin{figure}[t]
\centering
\subfloat[]{\includegraphics[width=\columnwidth]{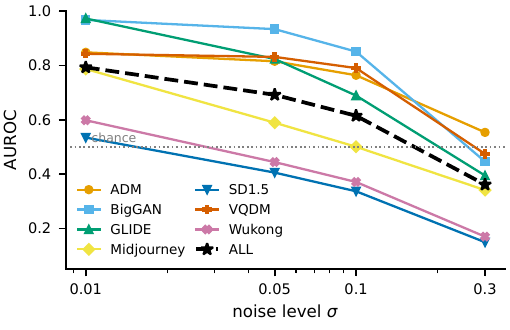}%
\label{fig:sigma}}\\[-1pt]
\subfloat[]{\includegraphics[width=\columnwidth]{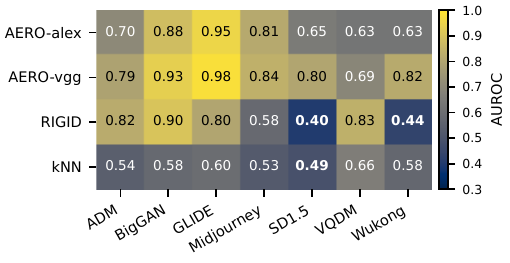}%
\label{fig:heat}}
\caption{Fragility visualization. (a)~RIGID-style AUROC vs.\ noise level
$\sigma$ (final-layer features, clean images, original pipeline): SD1.5 and
Wukong cross below the chance line as $\sigma$ grows, while BigGAN and GLIDE stay
high until $\sigma{=}0.3$. (b)~Method$\times$generator AUROC heatmap on the
bias-corrected \texttt{q95} condition (RIGID at $\sigma{=}0.05$, last; kNN at
$k{=}5$); bold white cells are below chance.}
\label{fig:fragility}
\end{figure}

Training-free scores come with a \emph{direction assumption}---e.g., ``real
images are more robust to perturbation''. Table~\ref{tab:rigidgrid} and
Fig.~\ref{fig:sigma} show this
assumption is contingent on hyperparameters in a generator-dependent way. At
$\sigma{=}0.05$, SD1.5 and Wukong (an SD-family model) are \emph{inverted}: their
fakes are more noise-robust than real images, so the score ranks them on the
wrong side. Reducing $\sigma$ to $0.01$ restores the assumed direction for all
seven generators and both depths, and is also globally optimal (ALL $=0.793$);
increasing $\sigma$ to $0.3$ with final features inverts six of the seven
generators. Feature depth interacts non-uniformly: mid-layer features are more
direction-stable for SD-family fakes but \emph{less} stable for VQDM ($0.480$ mid
vs.\ $0.831$ last at $\sigma{=}0.05$). This is consistent with concurrent reports
of exceptions to the robustness postulate~\cite{huang2026intermediate}, and goes
further: a deployed threshold tuned at one $\sigma$ can be \emph{worse than
random} on a generator family it was never validated on, without any distribution
shift in the usual sense.

\subsection{Anatomy of the inversion}
\label{ssec:anatomy}
\begin{table}[t]
\centering
\caption{Mean raw cosine similarity between original and noise-perturbed
DINOv2 final-layer embeddings (clean images, original pipeline). Bold: fake
mean \emph{exceeds} the real mean, i.e., fakes are more noise-robust and the
score direction inverts. The mean ordering predicts on which side of chance
the AUROC falls (Table~\ref{tab:rigidgrid}) in all 21 generator cells.}
\label{tab:meancos}
\footnotesize
\setlength{\tabcolsep}{8pt}
\begin{tabular}{lccc}
\toprule
 & $\sigma{=}0.01$ & $\sigma{=}0.05$ & $\sigma{=}0.3$ \\
\midrule
Real       & 0.9952 & 0.9465 & 0.3889 \\
\midrule
ADM        & 0.9812 & 0.8788 & 0.3514 \\
BigGAN     & 0.9623 & 0.8109 & \textbf{0.4184} \\
GLIDE      & 0.9615 & 0.8747 & \textbf{0.4623} \\
Midjourney & 0.9883 & 0.9375 & \textbf{0.4991} \\
SD1.5      & 0.9948 & \textbf{0.9568} & \textbf{0.6500} \\
VQDM       & 0.9779 & 0.8714 & \textbf{0.4011} \\
Wukong     & 0.9934 & \textbf{0.9531} & \textbf{0.6371} \\
\bottomrule
\end{tabular}
\end{table}

Because an inverted AUROC could also be produced by a trivial sign or labeling
bug, we verified that the inversion is a property of the data. First, the score
is a single raw cosine-similarity column computed by one formula for all
images; a sign error would flip \emph{all} generators uniformly and cannot
selectively flip SD1.5 and Wukong while leaving ADM--VQDM intact. Second, the
effect is visible before any AUROC is computed: Table~\ref{tab:meancos} reports
the mean raw similarity per generator. At $\sigma{=}0.05$ the SD-family fakes
($0.9568$, $0.9531$) sit \emph{above} the real mean ($0.9465$)---they genuinely
move less in feature space under noise---while the other five generators sit
below; the fake-vs-real mean ordering predicts which side of chance every one
of the 21 generator$\times\sigma$ AUROC cells falls on. At $\sigma{=}0.01$ all
seven fake means drop below the real mean, restoring the assumed direction,
though only marginally for SD1.5 ($0.9948$ vs.\ $0.9952$, AUROC $0.534$). At
$\sigma{=}0.3$ the separation reverses dramatically: SD1.5 retains $0.650$
mean similarity while real images collapse to $0.389$. Third, the phenomenon
is reproducible: two independent scoring runs (separate scripts, fresh noise
draws) correlate at Pearson $r{=}0.989$ over the 1{,}500 images, and an
independent pairwise Mann--Whitney implementation reproduces every AUROC. The
inversion is thus a real, $\sigma$-dependent property of SD-family images
under DINOv2 features, not an implementation artifact.

\subsection{Dataset format bias and the ``JPEG helps'' artifact}
\label{ssec:bias}
\begin{table}[t]
\centering
\caption{Bias-corrected pipeline: all images re-encoded at JPEG $q{=}95$ before
the $q{=}70/50$ degradation. The VGG variant now degrades monotonically; the
residual non-monotonicity of AlexNet localizes to BigGAN. Neither the 4-score nor
the 2-score fusion beats the best single score.}
\label{tab:bias}
\footnotesize
\begin{tabular}{lccc}
\toprule
Score & q95 & JPEG-70 & JPEG-50 \\
\midrule
AERO alex-512 & 0.749 & 0.753 & 0.760 \\
AERO vgg-512  & \textbf{0.835} & \textbf{0.790} & \textbf{0.775} \\
RIGID ($\sigma{=}0.05$, last) & 0.682 & 0.673 & 0.675 \\
kNN control   & 0.567 & 0.557 & 0.557 \\
\midrule
Fusion (z-score, all 4 scores) & 0.782 & 0.757 & 0.753 \\
Fusion (z-score, vgg$+$RIGID)  & 0.829 & 0.783 & 0.772 \\
\bottomrule
\end{tabular}
\end{table}

In the original pipeline, the AlexNet reconstruction AUROC \emph{increases}
under JPEG-50 (Table~\ref{tab:overall}), which would absurdly suggest that
compression helps detection. GenImage reals are JPEG-sourced while fakes are
PNG, so degradation initially harms reals less---a format confound, not a
detector property. After unified $q{=}95$ re-encoding (Table~\ref{tab:bias})
the VGG variant stays monotone ($0.835\to0.790\to0.775$), as expected, while
the AlexNet variant retains a
slight upward trend ($+0.011$, down from $+0.022$); per-generator analysis shows
it is driven almost entirely by BigGAN ($0.879\to0.979$ from q95 to JPEG-50),
whose low-resolution upsampled images interact pathologically with
compression---a finding we flag for benchmark designers rather than a bug we can
correct away.

\begin{table}[t]
\centering
\caption{Per-generator AUROC on the bias-corrected \texttt{q95} condition.
Bold: below chance. The heterogeneity of Table~\ref{tab:pergen} survives bias
correction: rankings, complementarity, and the SD-family inversion are
unchanged, confirming they are not format artifacts.}
\label{tab:pergen95}
\footnotesize
\setlength{\tabcolsep}{6.5pt}
\begin{tabular}{lcccc}
\toprule
Generator & alex-512 & vgg-512 & RIGID & kNN \\
\midrule
ADM        & 0.702 & 0.787 & 0.822 & 0.545 \\
BigGAN     & 0.879 & 0.934 & 0.901 & 0.579 \\
GLIDE      & 0.952 & 0.980 & 0.798 & 0.598 \\
Midjourney & 0.812 & 0.839 & 0.582 & 0.526 \\
SD1.5      & 0.648 & 0.798 & \textbf{0.400} & \textbf{0.490} \\
VQDM       & 0.626 & 0.690 & 0.834 & 0.659 \\
Wukong     & 0.628 & 0.816 & \textbf{0.436} & 0.575 \\
\bottomrule
\end{tabular}
\end{table}

Two further observations qualify the role of bias correction. First, it barely
moves the \emph{clean-level} numbers: from the original clean condition to
\texttt{q95}, overall AUROC changes by $+0.009$ (AlexNet), $+0.009$ (VGG),
$-0.010$ (RIGID), and $-0.000$ (kNN). What it changes is the compression
\emph{trend}---exactly the quantity robustness claims are based on. Second,
the per-generator structure survives correction
(Table~\ref{tab:pergen95}, Fig.~\ref{fig:heat}): reconstruction scores still detect GLIDE best and
VQDM worst, RIGID still excels on ADM, BigGAN, and VQDM and stays inverted on
SD1.5 ($0.400$) and Wukong ($0.436$), and the kNN control stays near chance.
The BigGAN compression pathology, by contrast, is shared by both
reconstruction backbones (from q95 to JPEG-50: $0.879\to0.979$ for AlexNet and
$0.934\to0.984$ for VGG) but barely affects RIGID ($0.901\to0.912$)---it is a
property of how BigGAN's upsampled, low-resolution images interact with
re-compression in pixel space, masked in the VGG \emph{overall} row of
Table~\ref{tab:bias} by larger drops on other generators.

\subsection{Fusion}
\label{ssec:fusion}
Per-condition z-score averaging yields AUROC $0.782/0.757/0.753$ (q95/70/50)
over all four scores and $0.829/0.783/0.772$ for the strongest pair
(vgg$+$RIGID)---both below the best single score (Table~\ref{tab:bias}), despite
the complementary per-generator behavior in Table~\ref{tab:pergen}. The
direction reversals of Table~\ref{tab:rigidgrid} explain why: averaging adds a
score whose sign is wrong on SD-family fakes. Exploiting complementarity thus
requires direction-aware (e.g., per-family sign-calibrated) combination, which we
leave to future work.

\section{Discussion}
\label{sec:discussion}

\textbf{Why are generators so heterogeneous?} The per-generator profiles in
Tables~\ref{tab:pergen} and~\ref{tab:pergen95} split the seven generators into
two camps. RIGID-style perturbation sensitivity is strongest on ADM, BigGAN,
and VQDM ($0.816$--$0.931$ clean)---pixel-space diffusion, GAN, and
vector-quantized models for which pronounced upsampling and spectral artifacts
are documented~\cite{corvi2023intriguing,tan2024npr}; small additive noise
plausibly disrupts exactly such high-frequency fingerprints, making fakes less
stable in feature space. The SD-family latent models (SD1.5, Wukong) behave
oppositely: their VAE-decoded outputs are locally smooth, and
Table~\ref{tab:meancos} shows their embeddings are \emph{more} noise-robust
than real ImageNet photographs once $\sigma$ exceeds the scale of natural
sensor noise---a reading consistent with, though not proven by, our
measurements. Reconstruction scores follow a different axis: they are best on
GLIDE and Midjourney and weakest on ADM and VQDM, and notably the SD1.5 fakes,
which share an autoencoder family with the scoring VAE, are \emph{not} the
easiest to detect ($0.647$ AlexNet clean); the expectation that
same-autoencoder fakes reconstruct best is at most partially supported, and
the backbone choice alone moves the SD1.5 cell by $+0.16$
(Table~\ref{tab:pergen}).

\textbf{What should evaluations report?} Our results suggest pooled AUROC is
close to meaningless for training-free scores: the pooled RIGID number
($0.692$) averages over per-generator values ranging from $0.408$ to $0.931$.
We recommend that audits and leaderboards (i)~report per-generator results,
with pooled numbers only as a summary; (ii)~validate the \emph{sign} of every
score on each generator family, e.g., via a $\sigma$- or strength-grid as in
Table~\ref{tab:rigidgrid}, before any threshold is deployed; (iii)~apply
provenance-controlled re-encoding by default and report both pipelines when
they disagree; and (iv)~report sufficient per-generator and per-condition breakdowns so that conclusions can be checked under different protocols.

\section{Audit conclusions}
\label{sec:conclusion}

\textbf{C1: Comparisons must control preprocessing and metric implementation.}
Backbone choice within the \emph{same} method ($+0.085$ AUROC) exceeds many
reported method-over-method gaps; preprocessing flips per-generator rankings by
up to $0.38$ AUROC. Papers should report backbone, resolution policy, and crop
policy as first-class experimental variables.

\textbf{C2: Direction assumptions are fragile and must be validated per
generator family.} A training-free score is not a detector until its sign is
verified across the generator families it will face; we show a standard
configuration that is worse than random on SD-family fakes while excellent on
BigGAN, and that na\"ive fusion inherits this failure (Sec.~\ref{ssec:fusion}).

\textbf{C3: Robustness numbers require provenance-controlled re-encoding.}
The ``compression helps'' artifact---present for the AlexNet reconstruction
score---shrinks and localizes to a single generator (BigGAN) once real and
fake share encoding provenance; benchmarks should ship bias-corrected variants
by default~\cite{grommelt2024fakeorjpeg,guillaro2025bfree}.

\textbf{Limitations.} The audit covers one benchmark family (GenImage with
ImageNet reals), seven generators, two compression levels, and three score
families at moderate scale (1{,}500 test images); conclusions about absolute
performance should not be extrapolated to other real-image distributions, and
per-generator cells rest on 100 fakes each, so small differences between
adjacent cells should not be over-read. Robustness is probed with JPEG
re-compression only; resizing, blurring, and social-network laundering
pipelines, as well as adversarial post-processing, are out of scope. The
perturbation score uses a single backbone (DINOv2-base) and the reconstruction
score a single VAE, so we cannot separate what is specific to these models
from what is generic to their families; for Midjourney and Wukong, whose
architectures are not fully public, the mechanistic readings of
Sec.~\ref{sec:discussion} remain hypotheses. ADM/VQDM remain weak for
reconstruction scores, and the kNN control uses a single configuration. The
RIGID $\sigma$-grid was run on clean images only, and all results are
threshold-free AUROC---we do not study threshold transfer or calibration,
which deployment would additionally require.
Extending the audit to newer generators (SDXL-~\cite{podell2024sdxl} and
FLUX-class), newer training-free scores (HFI~\cite{choi2024hfi},
EIRES~\cite{jiang2025eires}, ZED~\cite{cozzolino2024zed},
MINDER~\cite{tsai2024minder}), and further corruption types is ongoing.

\vfill\pagebreak

\begingroup
\fontsize{8.3}{9.6}\selectfont
\setlength{\itemsep}{1pt}
\bibliographystyle{IEEEbib}
\bibliography{refs}
\endgroup

\end{document}